\documentclass[a4paper, conference]{IEEEtran}
\usepackage{cite}
\ifCLASSINFOpdf
   \usepackage[pdftex]{graphicx}
\else
   \usepackage[dvips]{graphicx}
\fi
\usepackage{amsmath}
\usepackage{algorithmic}
\usepackage{array}
\usepackage{url}
\usepackage{caption}
\usepackage{subcaption}
\usepackage{comment}
\usepackage{tabularx,booktabs}
\usepackage{amsmath,amssymb}
\newcolumntype{L}[1]{>{\raggedright\arraybackslash}m{#1}}
\newcolumntype{C}[1]{>{\centering\arraybackslash}p{#1}}

\begin{document}
\title{Exploring Severe Occlusion: Multi-Person 3D Pose Estimation with Gated Convolution}


% \author{\IEEEauthorblockN{Renshu Gu}
% \IEEEauthorblockA{University of Washington\\
% Seattle, 98195, USA\\
% Email: renshugu@uw.edu}
% \and
% \IEEEauthorblockN{Gaoang Wang}
% \IEEEauthorblockA{University of Washington\\
% Seattle, 98195, USA\\
% Email: gaoang.wang1991@gmail.com}
% \and
% \IEEEauthorblockN{Jenq-Neng Hwang}
% \IEEEauthorblockA{University of Washington\\
% Seattle, 98195, USA\\
% Email: hwang@uw.edu}}

\author{\IEEEauthorblockN{Renshu Gu$^{1,3}$, Gaoang Wang$^{2,*}$, Jenq-Neng Hwang$^3$}
\IEEEauthorblockA{$^1$Hangzhou Dianzi University, Hangzhou, Zhejiang 310018, China\\
$^2$Zhejiang University-University of Illinois at Urbana-Champaign Institute, Haining, Zhejiang 314400, China\\
$^3$University of Washington, Seattle, 98195, USA\\
$^{*}$Email: gaoangwang@intl.zju.edu.cn}
}


%Telephone: (800) 555--1212, Fax: (888) 555--1212}


\maketitle

\begin{abstract}
3D human pose estimation (HPE) is crucial in many fields, such as human behavior analysis, augmented reality/virtual reality (AR/VR) applications, and self-driving industry. Videos that contain multiple potentially occluded people captured from freely moving monocular cameras are very common in real-world scenarios, while 3D HPE for such scenarios is quite challenging, partially because there is a lack of such data with accurate 3D ground truth labels in existing datasets. In this paper, we propose a temporal regression network with a gated convolution module to transform 2D joints to 3D and recover the missing occluded joints in the meantime. A simple yet effective localization approach is further conducted to transform the normalized pose to the global trajectory. To verify the effectiveness of our approach, we also collect a new moving camera multi-human (MMHuman) dataset that includes multiple people with heavy occlusion captured by moving cameras. The 3D ground truth joints are provided by accurate motion capture (MoCap) system. From the experiments on static-camera based Human3.6M data and our own collected moving-camera based data, we show that our proposed method outperforms most state-of-the-art 2D-to-3D pose estimation methods, especially for the scenarios with heavy occlusions. 
\end{abstract}


% \fi
\IEEEpeerreviewmaketitle

\section{Introduction}
% This demo file is intended to serve as a ``starter file''
% for IEEE conference papers produced under \LaTeX\ using
% IEEEtran.cls version 1.8b and later.
Human pose estimation (HPE) is gaining high attention in recent years, as it is crucial in many areas such as human behavior analysis, augmented reality/virtual reality (AR/VR) applications, and self-driving industry. There has been a great amount of reports on 3D human pose estimation for videos recorded by static cameras in the experimental setting. However, there exist limited reports of quantitative results for natural videos that are captured by moving cameras and contain multi-person interactions. 

Occlusion handling, including self-occlusion or inter-person occlusion, is an important issue in human pose estimation. Unfortunately, there is a lack of data with accurate ground truth for multi-person moving camera videos with inter-person occlusions. Multi-person 3D pose estimation with heavy occlusion is still not well-evaluated nor well-addressed in the literature.

In this paper, we aim to address and evaluate the occlusion problem in 3D human pose estimation. We first introduce a new dataset that has commonly encountered multi-person interactions, with various occlusion scenarios captured by a moving camera with ground truth being collected by a motion capture (MoCap) system simultaneously. The dataset includes several commonly seen actions. To estimate 3D pose from 2D pose, we propose a temporal regression network using gated convolutions, which is inspired by the image inpainting application \cite{yu2019free}. The estimated 3D position of each joint is with respect to the root position, and a global trajectory recovery approach is adopted to localize the joints based on the 3D  position in the camera coordinate. This simple yet effective global trajectory recovery approach is proposed based on 3D-to-2D back-projection and temporal continuity. 

In summary, we claim the following contributions:

1) To our best knowledge, we are the first to use temporal gated convolutions to recover missing poses and address the occlusion issues in the pose estimation. The idea of missing pose recovery is inspired by the image inpainting tasks. 

2) We present a new human pose dataset, which is aimed at multi-person 3D pose estimation with occlusion handling in moving camera environments.

3) A simple yet effective approach is proposed to transform normalized poses to the global trajectory into the camera coordinate.

4) We provide extensive experiments on videos containing multiple people from moving cameras, which are lacking in the current literature yet critical in today’s applications. Our solution provides great new opportunities to understand and predict human behaviors using freely moving cameras.

The organization of the paper is as follows: In Section \ref{sec:Related}, we review some related works of 3D human pose estimation based on monocular cameras. The data collection and the new dataset description are then introduced in Section \ref{sec:dataset}. In Section \ref{sec:approach}, the proposed approach of 3D human pose estimation with temporal gated convolution and global trajectory estimation is discussed in detail. Experimental setup and results are presented in Section \ref{sec:experiment} and Section \ref{sec:result}, respectively, followed by the conclusion in Section \ref{sec:conclusion}.

\section{Related Work}
\label{sec:Related}
Recent efforts of 3D human pose estimation, mainly CNN based, show promising results. However, severe or full occlusion remains a challenging issue, especially in multi-person pose estimation. Since the multi-person 3D pose dataset is rare in the literature, the inter-person occlusion problem is still not well-studied. 
For the remaining section, we will briefly review some related works of 3D human pose estimation and survey the related 3D human pose estimation datasets.

\subsection{3D Human Pose Estimation} \label{related3dHPE}

\textbf{One-stage 3D pose estimation}. This class of methods, mainly based on convolutional neural networks (CNNs), focuses on end-to-end reconstruction \cite{jiang20103d,li20143d,tekin2016direct,pavlakos2017coarse,zhou2017towards,rogez2017lcr,rogez2019lcr,mehta2018single,zanfir2018deep} by directly estimating 3D poses from RGB images without intermediate supervision. Since image cues are directly used, localizing and cropping the subject greatly affects the results. A ﬁne discretization of the 3D space around the subject is proposed in \cite{pavlakos2017coarse}, which employs a coarse-to-ﬁne prediction, given that input is a single RGB image. However, despite their attempt to evaluate on real-world scenarios, results are only reported on one sequence, and as they clarify, the available ground truth is  limited and not exactly accurate. A weakly-supervised transfer learning method that uses mixed 2D and 3D labels in a uniﬁed deep neural network is presented in \cite{zhou2017towards}.  Rogez et al. \cite{rogez2017lcr}, propose an end-to-end architecture LCR-net that contains a pose proposal generator, a classiﬁer that predicts the different anchor-pose labels, and an anchor-pose-speciﬁc regressor. They further extend LCR-net to LCR-net++ and adds additional synthetic training data and show better performance on partially occluded or truncated persons in \cite{rogez2019lcr}. However, LCR-Net and LCR-net++ do not use any temporal information, thus they cannot handle full-body or severe occlusions. Mehta et al.'s work \cite{mehta2018single} uses occlusion-robust pose-maps (ORPM) to enable pose inference under partial occlusions within a single frame. 

\textbf{Two-stage 3D pose estimation.} The second class of methods first predicts 2D joint positions in image space (keypoints) which are subsequently lifted to 3D \cite{martinez2017simple,chen20173d,tekin2017learning,rayat2018exploiting,gu2019efficient,gu2019multi}. Recent work \cite{martinez2017simple} shows that given accurate 2D keypoints, predicting 3D poses is in fact relatively straightforward.  \cite{jiang20103d,chen20173d} simply search for a predicted set of 2D keypoints over a large set of 2D keypoints to find the corresponding 3D pose. Some approaches \cite{tekin2017learning} fuse 2D joints, image cues, and 3D joint coordinates along the way. \cite{zhou2017towards} and \cite{moon2019camera} predicts the depth of 2D joints to obtain the 3D pose. The advantages of two-stage methods include that (1) 2D pose is easy to predict with a large amount of 2D pose estimation data, since 2D ground truth data is much cheaper to obtain compared to 3D ground truth, (2) 2d-to-3d methods are not sensitive to diverse scenarios and environments. However, the two-stage solutions are more sensitive to the 2D pose estimation performance. For challenging scenarios with severe occlusions in the videos, the unreliable 2D keypoints may have a large influence on the 3D estimation performance. How to obtain robust 3D results even with noisy and occluded 2D keypoints as the input remains a challenging task.
% \gaoang{what are the advantages and disadvantages? For example, advantage: 2d pose are easy to predict with large amount of 2d pose estimation data. 2d-to-3d methods are not sensitive to diverse scenarios and environments. Disadvantage: the 3d estimation performance largely depends on the 2d estimation performance. Sensitive to the 2d performance.}

% Early 3D pose estimation approaches also focus on estimating poses for a single person, without considering multi-person scenarios. Recently, more efforts \cite{rogez2017lcr,rogez2019lcr,mehta2018single} have been reported for multi-person 3D pose estimation. Occlusion-robust pose-maps \cite{mehta2018single}, which is trained on a newly introduced dataset MuCo-3DHP\cite{mehta2018single}, is proposed to encode the 3D joint locations of all people in the scene. 

\textbf{3D pose estimation with sequential input.} In recent years, researchers try to utilize temporal information to estimate the 3D pose with sequential input. Meanwhile, the occlusion problem is gaining more attention \cite{rayat2018exploiting,pavllo20193d,cheng2019occlusion,lin2019trajectory}. A sequence-to-sequence
network composed of layer-normalized LSTM units is proposed in \cite{rayat2018exploiting}, which imposes temporal smoothness constraints during training. Dilated temporal convolutions are employed to capture long-term temporal information in \cite{pavllo20193d}. An occlusion-aware network for video using incomplete 2D keypoints, instead of complete but incorrect 2D keypoints, is proposed in \cite{cheng2019occlusion}. However, most of the existing works focus on self-occlusion with single person pose estimation. Moreover, the recovered poses are compared with respect to the root position in most experiments. However, poses in the camera coordinate are far more needed in practical applications. Different from existing works, our proposed method handles inter-person occlusions and long-time occlusions, which is quite common in real-world scenarios. Furthermore, our framework can efficiently estimate global trajectories and each subject's relative positions independent of any dataset. 

\textbf{3D human pose estimation datasets.} The most commonly used datasets for 3D human pose estimation include Human3.6M \cite{ionescu2013human3}, HumanEva \cite{sigal2006humaneva}, TotalCapture  \cite{xiang2019monocular}, MPI-INF-3DHP \cite{mehta2017monocular}, MuPoTS-3D\cite{mehta2018single} and 3DPW \cite{von2018recovering}. Human3.6M is a large-scale dataset featuring a single person at the center of videos recorded by a commercialized MoCap system. MoCap system is considered the most accurate system for obtaining 3D ground truth in the literature. However, the cameras in the MoCap system are static. HumanEva is similarly captured by a MoCap system much earlier and contains simpler poses. TotalCapture captures diverse body motion using a massively multi-view sensor system, where the sensors are also statically installed.   MPI-INF-3DHP tries to include more challenging poses in the wild, but still focuses on single persons and static cameras. MuPoTS-3D is a multi-person dataset recorded by a static camera.  3DPW is the first dataset that includes video footage taken from a moving phone camera. However, it uses IMUs to record ground truth, which is reported to possess a 26mm average 3D pose error when verified with the Mocap system. 
% Our new dataset, on the other hand, utilizes the accurate Mocap system while providing moving camera footage, and also includes common action classes that are not available in 3DPW. 

\section{The Moving Camera Multi-Human Dataset}
\label{sec:dataset}
Unlike existing 3D human pose estimation datasets, we propose a new dataset, multi-person moving camera (MMHuman) dataset, which is aimed at providing accurate MoCap ground truth for multi-person interaction videos captured by moving cameras, and intentionally includes occlusions and human interactions that are common in real-world scenarios. 

\subsection{Data Collection and Design}
\begin{figure}[t]
\begin{center}
\includegraphics[width=0.8\linewidth]{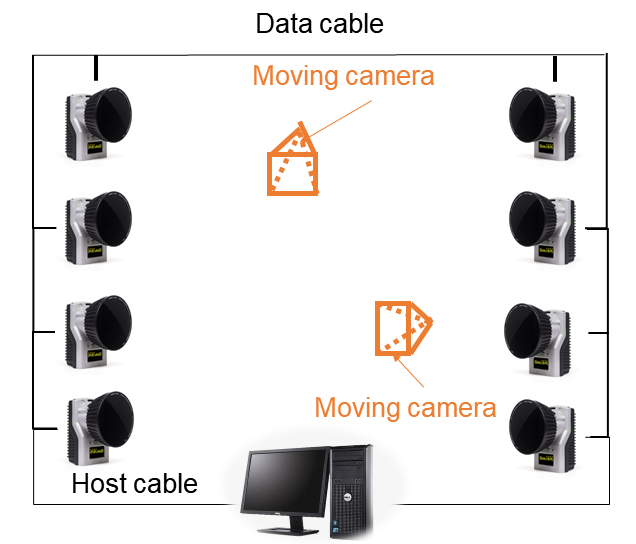}
\end{center}
   \caption{The motion capture environment.}
\label{fig:MoCap} 
\end{figure}

MMHuman dataset was captured via a commercialized MoCap system and additional mobile cameras. The MoCap system consists of 8 static IR cameras that record synchronized high-resolution 120Hz videos. The additional moving cameras are hand-held commercial smartphones, manually synchronized with the MoCap system. The MoCap system relies on small reﬂective markers attached to the subjects' bodies and tracks them over time. Tracking maintains the labeled identity and propagates it through time from an initial pose which is labeled either manually or automatically. A ﬁtting process uses the position and identity of each of the body labels to infer accurate pose parameters. The final 3D ground truth pose is provided in the global coordinate, i.e. the calibrated MoCap coordinate. Our MoCap environment is shown in Fig.~\ref{fig:MoCap}. 

We do not see our proposed dataset as an alternative to existing datasets; rather, MMHuman complements existing ones with accurate ground truth of new multi-person interaction sequences shot by freely moving cameras. Our new dataset allows a quantitative evaluation of state-of-the-art approaches for multi-human interaction cases that have heavy and inter-person occlusions.

\subsection{Dataset Description}
\begin{figure}[t]
\begin{center}
\begin{subfigure}{0.4\linewidth}
    \centering
    \includegraphics[width=\linewidth]{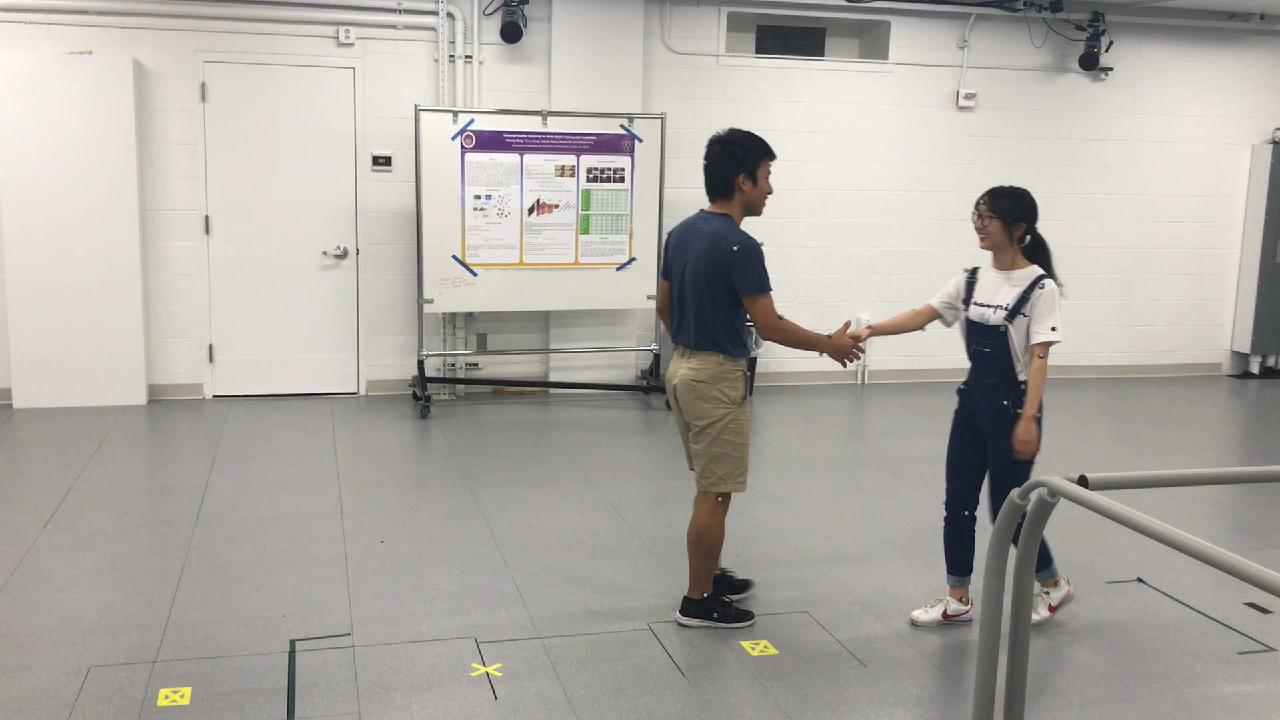}
    \caption{ShakeHand}
\end{subfigure}
\begin{subfigure}{0.4\linewidth}
    \centering
    \includegraphics[width=\linewidth]{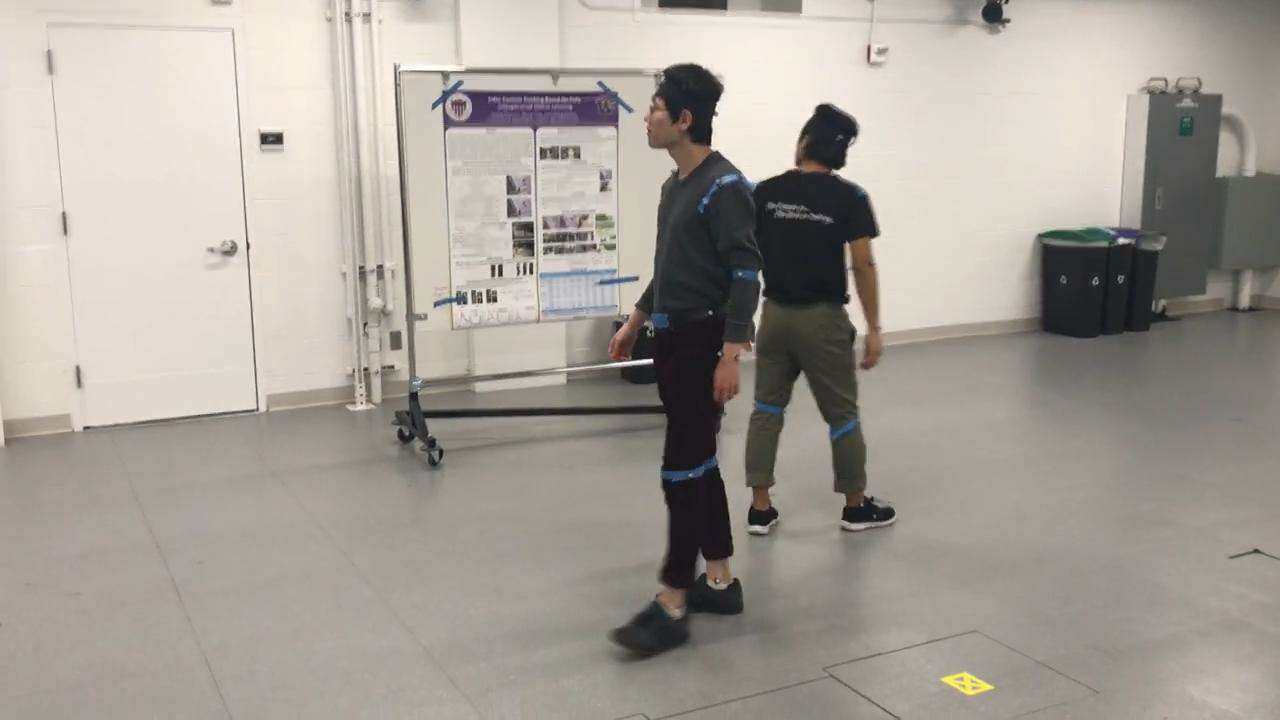}
    \caption{WalkCross}
\end{subfigure}
\begin{subfigure}{0.4\linewidth}
    \centering
    \includegraphics[width=\linewidth]{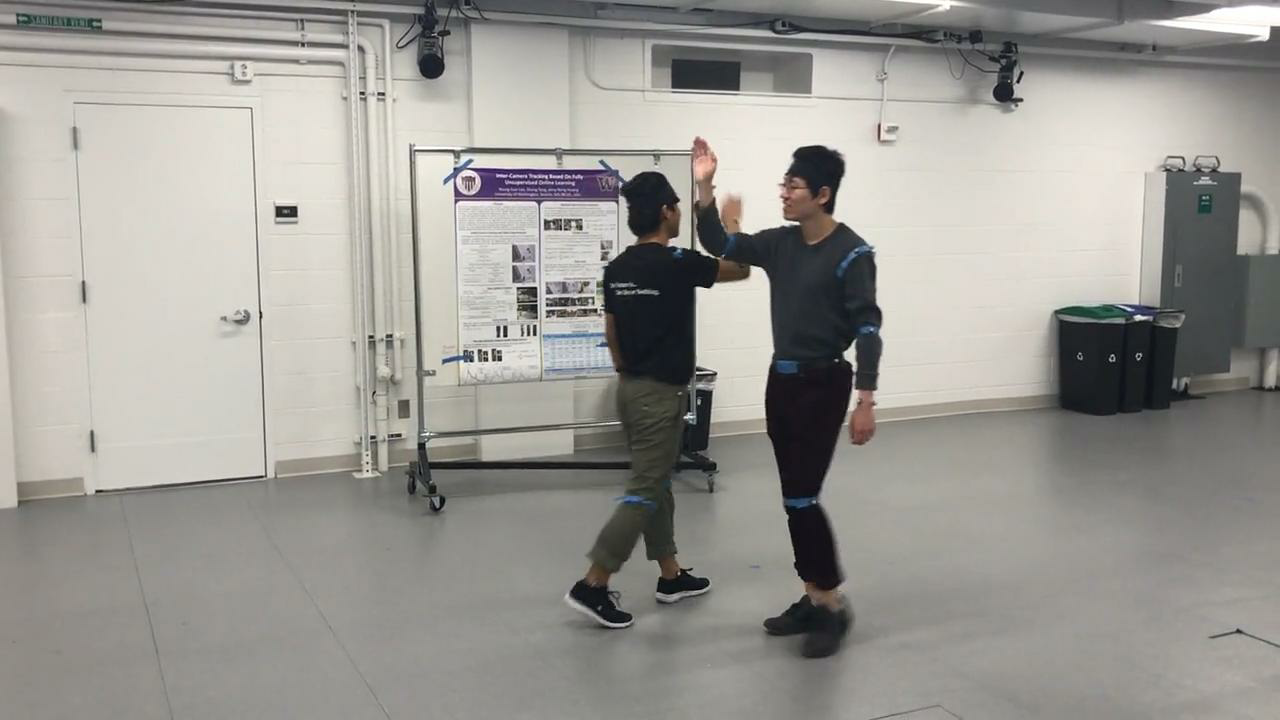}
    \caption{HighFive}
\end{subfigure}
\begin{subfigure}{0.4\linewidth}
    \centering
    \includegraphics[width=\linewidth]{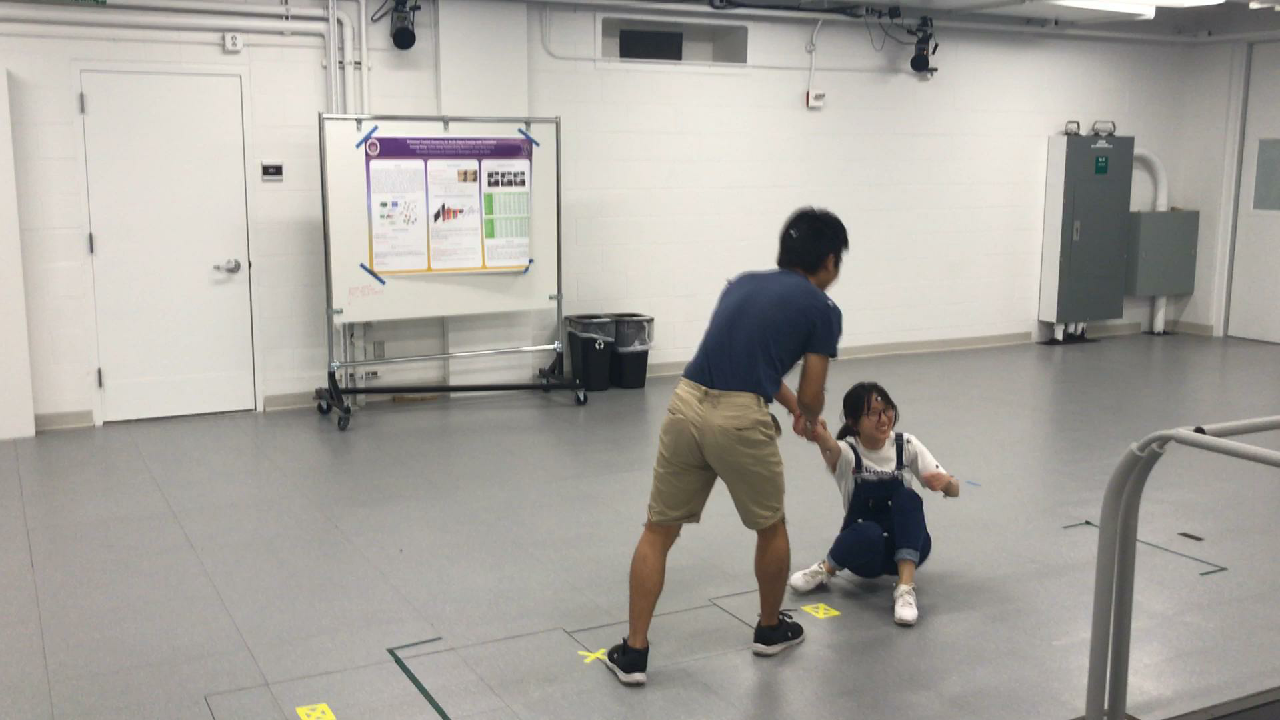}
    \caption{PullUp}
\end{subfigure}
\begin{subfigure}{0.4\linewidth}
    \centering
    \includegraphics[width=\linewidth]{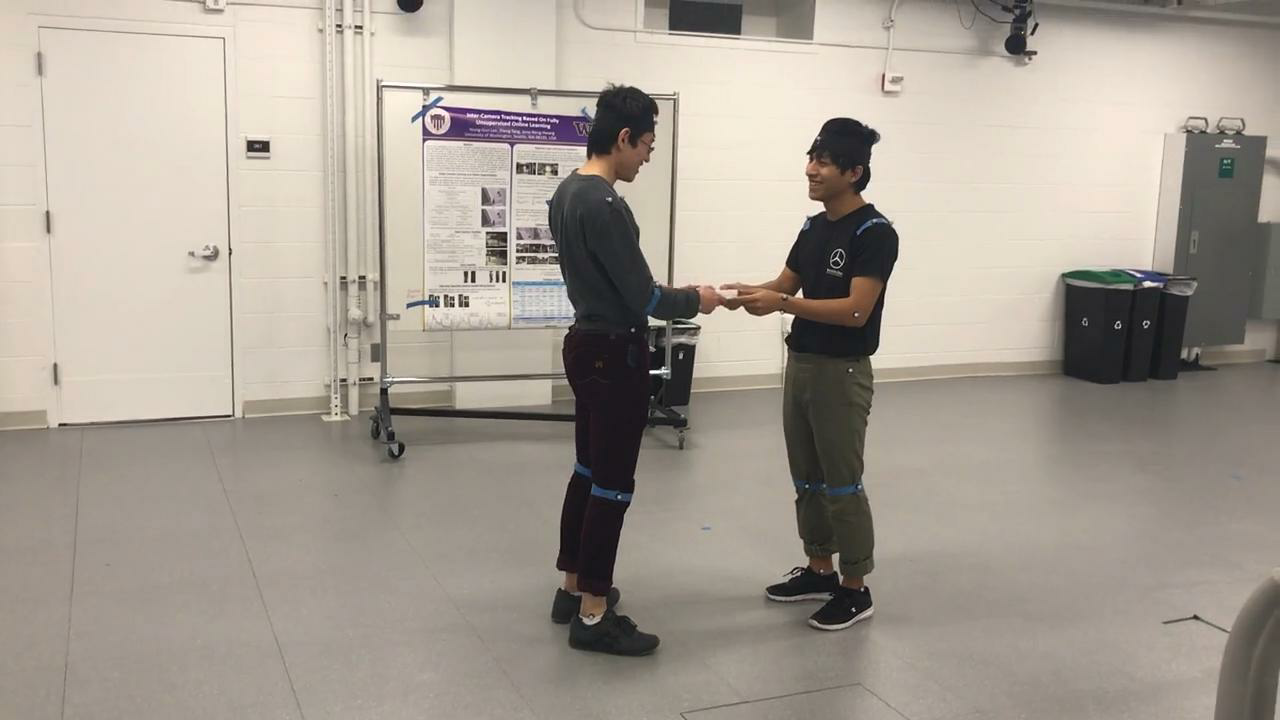}
    \caption{HandOver}
\end{subfigure}
\begin{subfigure}{0.4\linewidth}
    \centering
    \includegraphics[width=\linewidth]{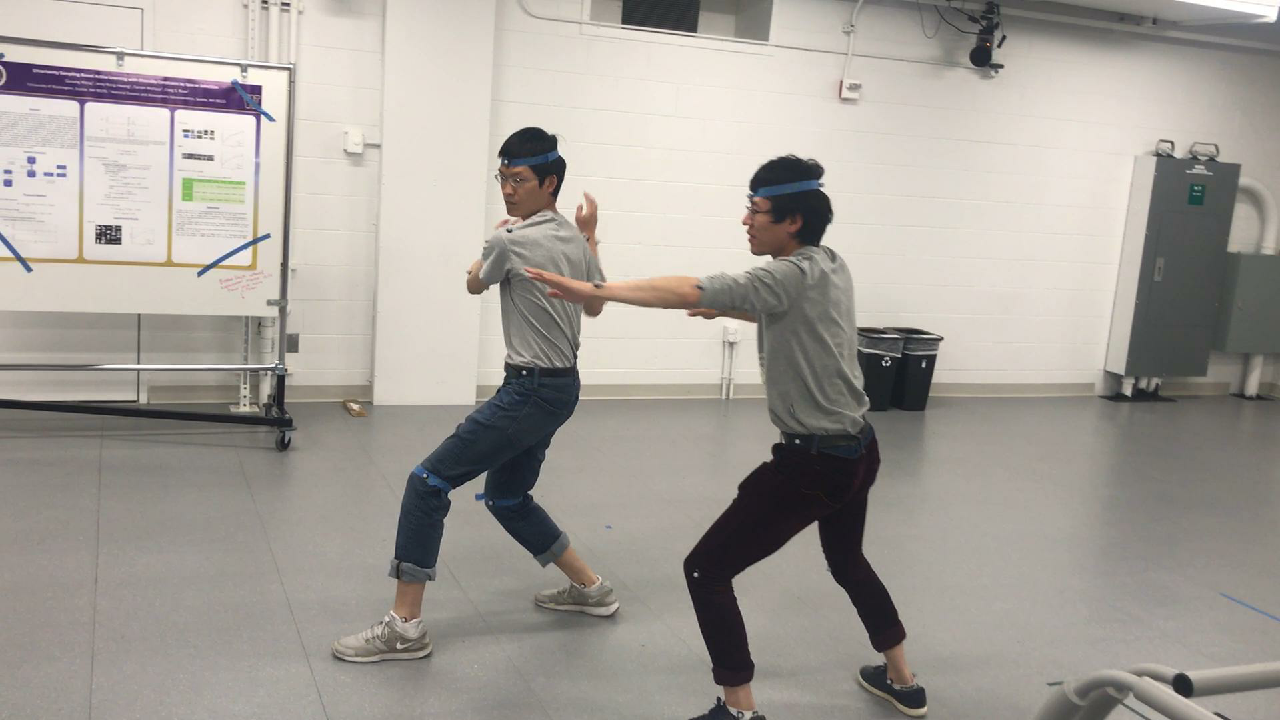}
    \caption{Kungfu}
\end{subfigure}
\end{center}
\caption{Examples of actions in MMHuman.}
\label{fig:mmhuman} 
\end{figure}

The dataset consists of 6 actions that involve common human-interactions and other variations, as shown in Fig.~\ref{fig:mmhuman}. Table \ref{tab:summmhuman} shows a summary of our dataset. Summing up all videos recorded by different cameras, the total number of frames available is also provided. 
% \gaoang{Add one example figure for each action.}

% \begin{table}[h]

\begin{table}[!t]
\vspace{1em}
\captionsetup{font=footnotesize, margin=0em}
\caption{MMHuman Dataset Description}
\label{tab:summmhuman}
\begin{center}
\begin{tabular}{m{1.2cm}C{0.7cm}C{0.7cm}C{0.7cm}C{0.7cm}C{0.7cm}C{0.7cm}}
\hline\noalign{\smallskip}
& Shake Hand & Walk Cross & High Five & Pull Up & Hand Over & Kungfu \\
\noalign{\smallskip}
\hline
\noalign{\smallskip}
{\#} Videos & 6 & 6 & 5 & 3 & 5 & 2\\
\hline
\noalign{\smallskip}
{\#} Frames & 11,573 & 11,379 & 9,537 & 5,809 & 9,570 & 3,883\\
\hline
Total Frames & \multicolumn{6}{c}{51,751}\\
\hline
\end{tabular}
\end{center}
\end{table}

\section{Approach}
\label{sec:approach}
In this section, we explain our proposed 3D pose estimation approach in detail, including the temporal regression model with modified gated convolution, model architecture, and global pose trajectory estimation.

\subsection{Temporal Gated Convolution}

\begin{figure*}[t]
\begin{center}
\includegraphics[width=0.5\linewidth]{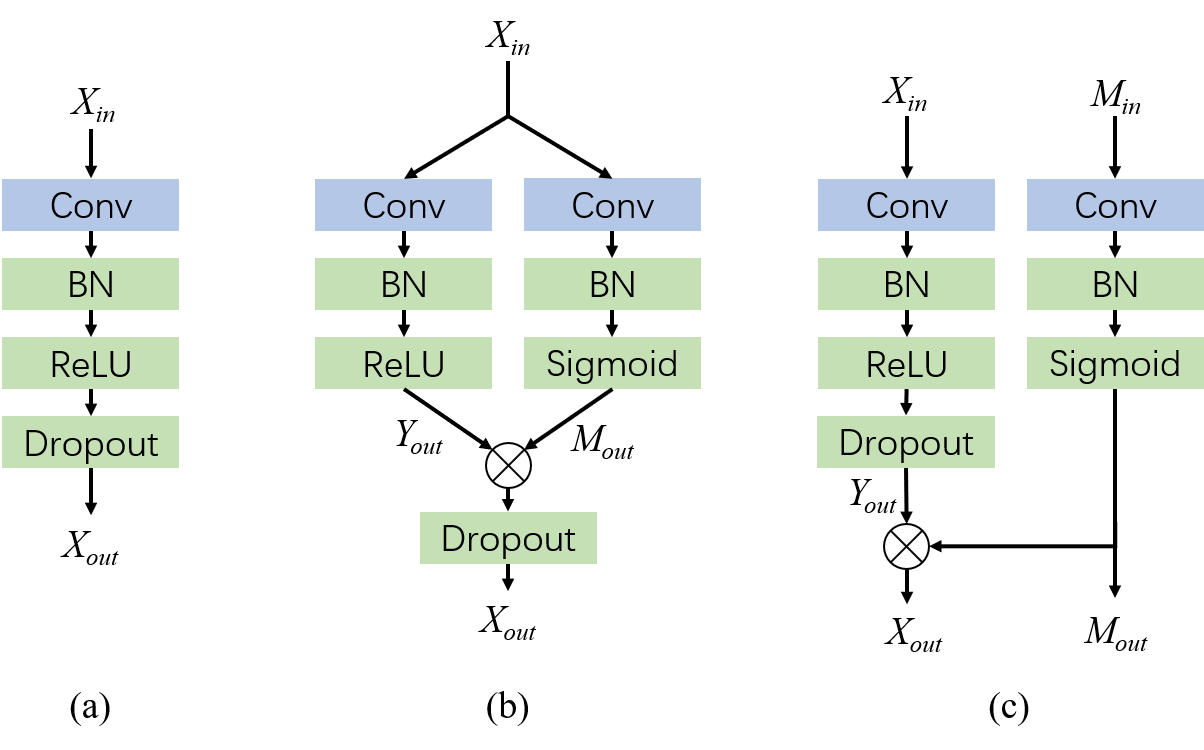}
\end{center}
   \caption{Temporal convolution layer in the pose estimation model. (a): The plain convolution layer used in \cite{pavllo20193d}. (b): The soft gated convolution proposed in \cite{yu2019free} for inpainting task. (c): Modified two-stream soft gated convolution proposed in this paper.}
\label{fig:model} 
\end{figure*}

Given a temporal sequence of detected 2D joint locations $\boldsymbol{x} \in \mathbb{R}^{2N_{jt} \times T}$, our goal is to reconstruct the 3D pose $\boldsymbol{X} \in \mathbb{R}^{3N_{jt} \times 1}$ of the center frame, where $T$ is the temporal window size, and $N_{jt}$ is the number of joints in the pose model. 

When occlusions occur, the 2D joints estimated by human pose detectors are either missing or with low confidence. Such noisy 2D joints adversely affect the accuracy of the 3D pose estimation. Intuitively, we can use attention or soft gated convolution to focus on non-occluded joints in the regression model, which is a well-known technique in image inpainting networks, such as \cite{liu2018image,yu2018generative,yu2019free}. Similarly, we can also treat 3D pose estimation as a special case of inpainting task when occlusions occur. For the initial input, additional to feed in the sequential 2D poses, we also feed the binary occlusion mask, $\boldsymbol{M} \in \mathbb{R}^{2N_{jt} \times T}$. Two separate convolution kernels are adopted to obtain the feature maps and soft masks individually. The soft masks can be regarded as soft gating operation or attention on the output features. Specifically, the soft gated convolution is defined as follows,
\begin{equation}
\begin{aligned}
    & \boldsymbol{Y}^{l} = \phi \left(\boldsymbol{X}^{l-1} \ast \boldsymbol{W}_f^{l} \right), \\
    & \boldsymbol{M}^{l} = \sigma \left( \boldsymbol{X}^{l-1} \ast \boldsymbol{W}_g^{l} \right), \\
    & \boldsymbol{X}^{l} = \boldsymbol{Y}^{l} \otimes \boldsymbol{M}^{l},
\label{eq:gated_conv}
\end{aligned}
\end{equation}
where $\boldsymbol{X}^{l-1}$ is the output of the previous layer, $\boldsymbol{W}_f^{l}$ are the convolution kernels of the feature map, $\boldsymbol{W}_g^{l}$ are the convolution kernels of the soft mask, $\boldsymbol{Y}^{l}$ is the feature map of the current layer, $\boldsymbol{M}^{l}$ is the mask of the soft gate, $\boldsymbol{X}^{l}$ is the output of the current layer, $\otimes$ represents element-wise multiplication, and $\phi$ and $\sigma$ are the activation functions related to feature map and soft mask, respectively. The gated convolution is illustrated in Fig.~\ref{fig:model}(b).

However, in our experiments, the soft gated convolution defined in Eq.~(\ref{eq:gated_conv}) does not perform very well. This is reasonable since the soft mask and feature maps are calculated from the same input. It is difficult for the networks to learn different roles of soft gating and feature map extraction at the same time without further supervision. Since the soft gating is more correlated to the input masks rather than the input feature maps, we separate the soft gating and feature map extraction into two streams as shown in Fig.~\ref{fig:model}(c). Specifically, we modify Eq.~(\ref{eq:gated_conv}) as follows,
\begin{equation}
\begin{aligned}
    & \boldsymbol{Y}^{l} = \phi \left(\boldsymbol{X}^{l-1} \ast \boldsymbol{W}_f^{l} \right), \\
    & \boldsymbol{M}^{l} = \sigma \left( \boldsymbol{M}^{l-1} \ast \boldsymbol{W}_g^{l} \right), \\
    & \boldsymbol{X}^{l} = \boldsymbol{Y}^{l} \otimes \boldsymbol{M}^{l},
\label{eq:gated_conv2}
\end{aligned}
\end{equation}
where the soft mask is calculated only from the soft mask of the previous layer. This modification can make the network training benefit from easy convergence.

\subsection{Network Architecture}
We follow the same regression architecture used in \cite{pavllo20193d}. Apart from a sequence of 2D joints, we also input the occlusion mask. After the first temporal gated convolution layer, 4 skip blocks are used for regression to estimate the 3D joints. For each layer, we replace the plain convolution as shown in Fig.~\ref{fig:model}(a) with the modified temporal gated convolution as shown in Fig.~\ref{fig:model}(c). For each layer, we input both the feature maps and the soft mask from the previous layer. The temporal dilated convolution is conducted separately for input feature maps and soft mask with different kernels, followed by the batch normalization and activation. We use ReLU for feature map extraction and Sigmoid function for the soft gating. For the supervision, we use the mean square error between predicted 3D joints and ground truth as the loss, which is defined as,

\begin{equation}
    \mathcal{L} = \frac{1}{N_{jt}}\sum_{i=1}^{N_{jt}}||\hat{\boldsymbol{X}}_{i}-\boldsymbol{X}_{i}^{\ast}||^2,
\end{equation}
where $\hat{\boldsymbol{X}}$ is the predicted 3D joints and $\boldsymbol{X}^{\ast}$ is the 3D ground truth.

\subsection{Global Pose Trajectory Estimation via Back-Projection}
We adopt the same output format as \cite{pavllo20193d}, i.e. the output of the temporal regression network is the position of all joints with respect to the root location. In \cite{pavllo20193d}, to obtain the global position of each joint, a second network called trajectory model, is trained separately to estimate the trajectory of the human in the camera coordinate. However, this extra effort can be efficiently replaced by a simple optimization approach that utilizes the temporal consistency and 2D-3D back-projection relationship. 

For simplicity, we illustrate how to reconstruct the trajectory of one person in the following subsection. For multi-person scenarios, we can estimate all the individual trajectories separately. Denote the root position in the camera coordinate at frame $t$ as $\boldsymbol{C}_t$. Then we can define the projection error as follows,
\begin{equation}
    \epsilon_t = \sum_{i=1}^{N_{jt}}||\rho(\boldsymbol{X}_{i,t}+\boldsymbol{C}_{t})-\boldsymbol{x}_{i,t}||^2,
\end{equation}
where $\rho$ is the projection of the pinhole camera, which maps the 3D coordinate into image coordinate, i.e.,
\begin{equation}
\begin{aligned}
    & \rho_x(\boldsymbol{X}) = \frac{fX}{Z}+c_x, \\
    & \rho_y(\boldsymbol{X}) = \frac{fY}{Z}+c_y,
\end{aligned}
\end{equation}
where $\boldsymbol{X}=[X, Y, Z]^T$ represents the 3D point in the camera coordinate, $f$ is the focal length,  $\{c_x,c_y\}$ is the center of the image coordinate.

Notice that all the joints share the same global trajectory $\boldsymbol{C}_t$ at each frame. Then the root trajectory $\boldsymbol{C}_t$ can be estimated by minimizing the projection error, i.e.,
\begin{equation}
    \hat{\boldsymbol{C}}_t = \arg \min_{\boldsymbol{C}_t} \epsilon_t,
\end{equation}

Since the back-projection is only available for non-occluded joints, the occlusion mask is introduced to constrain the error only on non-occluded joints, i.e., 
\begin{equation}
\label{eq:single-frame}
    \epsilon_t = \sum_{i=1}^{N_{jt}}(1-\boldsymbol{M}_{i,t})||\rho(\boldsymbol{X}_{i,t}+\boldsymbol{C}_{t})-\boldsymbol{x}_{i,t}||^2,
\end{equation}
where $\boldsymbol{M}_{i,t}$ is defined as 
\begin{equation}
    \boldsymbol{M}_{i,t} = 
    \begin{cases}
    1, \ \text{if} \ i\ \text{is occluded},\\
    0, \ \text{if} \ i\ \text{is visible}.
    \end{cases}
\end{equation}

However, if the human is fully occluded in certain frames $t$, then the projection error $\epsilon_t$ becomes zero, and solving $\boldsymbol{C}_t$ is impossible. As a result, we adopt the temporal continuity of the person's movement and estimate multi-frame trajectory simultaneously as follows,
\begin{equation}
\begin{aligned}
    & \hat{\boldsymbol{C}} = \arg \min_{\boldsymbol{C}} \sum_{t=1}^{F}\epsilon_t + \lambda_1 \sum_{t=2}^{F} ||\boldsymbol{C}_t-\boldsymbol{C}_{t-1}||^2 \\
    & \ \ \ \ \ +\lambda_2 \sum_{t=2}^{F-1}||\boldsymbol{C}_{t-1}+\boldsymbol{C}_{t+1}-2\boldsymbol{C}_{t}||^2,
\end{aligned}
\label{eq:multi-frame}
\end{equation}
where $\lambda_1$ and $\lambda_2$ control the first order and second order temporal continuity, respectively, and $F$ is the number of frames used for estimation. From Eq.~(\ref{eq:multi-frame}), the global pose trajectory can be estimated even if full-body occlusions occur.

\section{Experimental Setup and Results}
\label{sec:experiment}

\subsection{Datasets}
We evaluate the performance of our proposed method on two motion capture datasets, Human3.6M, and our recorded dataset MMHuman. Following previous work \cite{pavlakos2017coarse,tekin2017learning,martinez2017simple,sun2017compositional,fang2018learning,pavlakos2018ordinal,yang20183d,luvizon20182d,pavllo20193d} on Human3.6M, we adopt a 17-joint skeleton, training on five subjects (S1, S5, S6, S7, S8), and testing on two subjects (S9 and S11). 

In our experiments, we consider two evaluation protocols: Protocol 1 is the mean per-joint position error after alignment with the ground truth in translation, rotation, and scale \cite{pavllo20193d,martinez2017simple,sun2017compositional,fang2018learning,pavlakos2018ordinal,yang20183d,rayat2018exploiting}. Protocol 2 aligns predicted poses with the ground-truth only in scale in the camera coordinate with the global trajectory. We do not use the mean per-joint position error without alignment as the evaluation metric since it is sensitive to the camera intrinsic parameters.

\subsection{Implementation Details for 2D Pose Estimation}
To fairly evaluate our performance, all the comparison methods use the same 2D inputs. For experiments on Human3.6M, following VideoPose3D \cite{pavllo20193d} we use fine-tuned cascaded pyramid network (CPN) keypoints \cite{chen2018cascaded}. For experiments on MMHuman, we use the pre-trained OpenPose \cite{cao2017realtime} to detect 2D keypoints of every frame. 

\subsection{Implementation Details for 3D Pose Estimation with Sequential 2D Joints}
\begin{figure}[t]
\begin{center}
\includegraphics[width=0.95\linewidth]{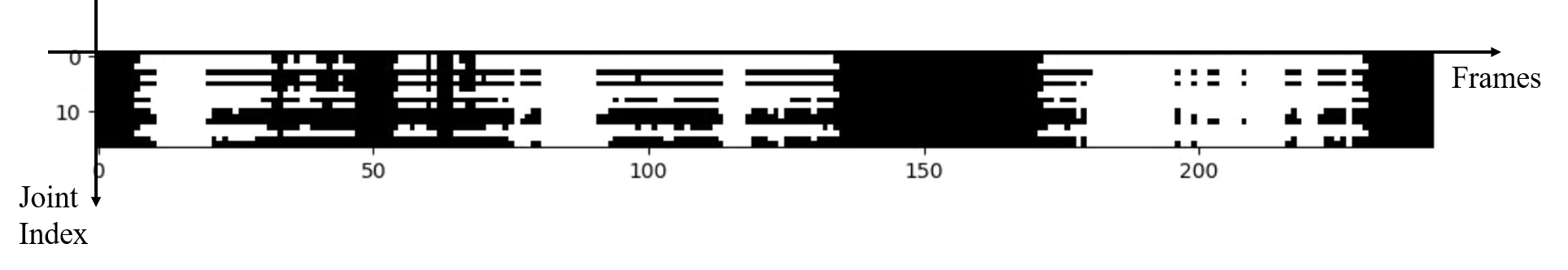}
\end{center}
   \caption{An example of the generated mask in 2D pose estimation. Horizontal and vertical axes represent frame index and joint index, respectively. The white color represents the occluded joints while the black color represents non-occluded joints.}
\label{fig:Mask} 
\end{figure}

For Human3.6M dataset, we randomly generate occlusion masks in both training and testing on the sequential 2D joints to test the effectiveness of temporal gated convolution on handling long-time occlusion. At first, we generate a matrix $P$ with the same size of the 2D input $\boldsymbol{x}$ with elements sampled from the uniform distribution $\mathcal{U}(0,1)$. Then the occlusion mask $\boldsymbol{M}$ can be set as $\boldsymbol{M}=\mathbb{I}_{\{P>\theta\}}$, where $\mathbb{I}$ is the indicator function that binarizes $P$ according to the threshold $\theta$. However, it is found that such settings cannot well represent long-time occlusion. The occlusion time span is usually short, and the 2D missing joints can be recovered well enough just by linear interpolation. Instead of using the uniformly sampled matrix for binarization directly, we convolve the matrix $P$ with the normalized all-one kernel $\boldsymbol{1}$ for each joint dimension separately to mimic long-time occlusion, i.e.,
\begin{equation}
    \boldsymbol{M} = \mathbb{I}_{\{\hat{P}>\theta\}},
\end{equation}
where
\begin{equation}
    \hat{P} = \frac{1}{K}P\ast \boldsymbol{1},
\end{equation}
where $\ast$ represents convolution operation and $K$ is the temporal kernel size. The larger the kernel size $K$ is, the stronger relationship across the temporal domain there is. As a result, we can change $K$ to represent the intensity of long-time occlusion. An example of the generated mask is shown in Fig.~\ref{fig:Mask}. We can see that the generated mask can have both partial occlusion and full occlusion.

\begin{table*}[!t]
\vspace{1em}
\captionsetup{font=footnotesize}
\begin{center}
\caption{3D Pose Error (Protocol 1, Errors in mm)}
\label{tab:prot1}
\begin{tabular}{L{3cm}C{0.9cm}C{0.9cm}C{0.9cm}C{0.9cm}C{0.9cm}C{0.9cm}C{0.9cm}C{0.9cm}}
\hline\noalign{\smallskip}
& Dir. & Disc. & Eat & Greet & Phone & Photo & Pose & Purch.\\
\noalign{\smallskip}
\hline
\noalign{\smallskip}
Hossain \textit{et al}. \cite{rayat2018exploiting} & 50.8 & 60.3 & 59.6 & 60.1 & 75.0 & 67.9 & 53.7 & 59.8 \\
Pavllo \textit{et al}. \cite{pavllo20193d} & 46.3 & 53.9 & 49.3 & 56.8 & 48.2 & 57.7 & 51.2 & 51.5 \\
Lin \textit{et al}. \cite{lin2019trajectory} & 51.3 & 52.5 & 49.7 & 55.0 & 47.1 & 64.9 & 47.8 & 50.2 \\
Proposed & \textbf{42.9} & \textbf{43.5} & \textbf{41.5} & \textbf{44.7} & \textbf{41.9} & \textbf{49.4} & \textbf{42.0} & \textbf{41.0}\\

\noalign{\smallskip}
\hline
\noalign{\smallskip}
& Sit & SitD. & Smoke & Wait & WalkD. & Walk & WalkT. & Avg.\\
\noalign{\smallskip}
\hline
\noalign{\smallskip}
Hossain \textit{et al}. \cite{rayat2018exploiting} & 71.1 & 84.7 & 62.5 & 59.2 & 68.1 & 69.9 & 72.0 & 65.0\\
Pavllo \textit{et al}. \cite{pavllo20193d} & 53.0 & 63.6 & 50.8 & 45.5 & 56.3 & 57.0 & 56.2 & 53.2\\
Lin \textit{et al}. \cite{lin2019trajectory} & 53.1 & 61.7 & 51.0 & 47.6 & 60.1 & 65.7 & 62.1 & 53.9\\
Proposed & \textbf{50.4} & \textbf{59.8} & \textbf{43.9} & \textbf{41.3} & \textbf{46.5} & \textbf{35.1} & \textbf{32.6} & \textbf{43.8}\\
\hline
\end{tabular}
\end{center}
\end{table*}

For MMHuman, since occlusions naturally occur in the video recording, we use the originally detected 2D keypoints without generating occlusion masks using the strategy mentioned above. If the confidence of the detected 2D keypoint is below a threshold (we set 0.3 in our experiments), then the joint is treated as under occlusion. Different from Human3.6M dataset, our MMHuman contains multi-person in the video, therefore we apply multi-object tracking as a pre-processing step for all the comparison methods to solve the association problem. We adopt the TrackletNet tracking (TNT) \cite{wang2019exploit} for human tracking since it is proven highly effective when handling complicated multi-object tracking cases with various occlusions for both static and moving cameras’ recording. The details of TNT can be found in \cite{wang2019exploit}. 

For the comparison approaches listed in the following section, missing joints from the inputs could cause trouble in the inference. Therefore, we use linear interpolation in the temporal domain to fill out the missing inputs instead of directly feeding incomplete data. This gives us the best performances we could possibly get using the comparison approaches.

\subsection{Hyper Parameter Setting}
For the temporal gated convolution model, we use Amsgrad \cite{reddi2019convergence} as the optimizer and train for 80 epochs. We set the batch size as 1024 in the training. For Human3.6M, we adopt an exponentially decaying learning rate schedule, starting from $\eta = 0.001$ with a shrink factor $\alpha = 0.95$ applied on each epoch.

For the global pose trajectory estimation, we set $\lambda_1=1$, $\lambda_2=1$ and use $F=100$ frames in the optimization.

\section{Experimental Results}
\label{sec:result}

\subsection{Temporal Gated Convolution Model}
We compare our method with recent state-of-the-art temporal based models \cite{rayat2018exploiting,pavllo20193d,lin2019trajectory}, where both open source codes with pre-trained models provided. To explore severe occlusion cases, we randomly generate binary masks with occlusion ratio equaling to 50$\%$ in our experiments, i.e., 50$\%$ input joints are treated as occluded joints. Table \ref{tab:prot1} shows the 3D pose error following protocol 1 on Human3.6M dataset. From the table, we can see our proposed method achieves promising results. It shows the effectiveness of using gated convolution to handle heavy occlusion in the 3D pose estimation.

To further analyze the impact of occlusions in the input 2D pose, We also report the pose reconstruction errors on Human3.6M under protocol 1 with different occlusion ratios in the generated mask in Table \ref{tab:err_occ}. The results meet our expectations. As the occlusion ratio increases, all methods have higher errors in the 3D pose estimation. When there is 50$\%$ percent of missing joints, our method achieves significantly better performance, reporting an average error of 43.8 mm, which is 9.4 mm better than the second-best. Moreover, our proposed method does not have a large fluctuation in the reconstruction errors given different settings of occlusion ratios. It shows the robustness of the proposed method facing different occlusion situations. Note that for the first column in the table when no occlusion occurs, the reconstruction error of the proposed method is slightly higher (about 0.7 mm) than VideoPose3D \cite{pavllo20193d}. This is because since we add a large number of challenging examples with occlusions during training, the data distribution is slightly different between the training set and the testing set if no occlusion occurs in the testing time. This will lead to a slightly higher error than \cite{pavllo20193d}. Besides that, when no occlusion occurs, the input mask becomes a zero-matrix, and the advantage of gated convolution is not well utilized. 

\begin{table}[t]
\vspace{1em}
\captionsetup{font=footnotesize}
\begin{center}
\caption{3D Pose Error on Human3.6M with Different Occlusion Ratios (Errors in mm)}
\label{tab:err_occ}
\begin{tabular}{L{2.5cm}C{1.2cm}C{1.2cm}C{1.2cm}}
\hline\noalign{\smallskip}
Occ. Ratio & 0$\%$ & 25$\%$ & 50$\%$ \\
\noalign{\smallskip}
\hline
\noalign{\smallskip}
Hossain \textit{et al}. \cite{rayat2018exploiting} & 44.1 & 57.2 & 65.0 \\
Pavllo \textit{et al}. \cite{pavllo20193d} & \textbf{36.5} & 40.3 & 53.2 \\
Lin \textit{et al}. \cite{lin2019trajectory} & 36.8 & 41.9 & 53.9 \\
Proposed & 37.2 & \textbf{39.2} & \textbf{43.8} \\
\hline
\end{tabular}
\end{center}
\end{table}

Table \ref{tab:prot2} shows the results with the occlusion ratio equaling to 50$\%$ following protocol 2 on Human3.6M dataset. The errors are compared in the camera coordinate. Since no global pose trajectory is provided for methods \cite{rayat2018exploiting,lin2019trajectory}, we only compare with \cite{pavllo20193d}. From the table, we can see that the proposed method also outperforms VideoPose3D \cite{pavllo20193d} which uses an additional network to obtain the global trajectory. It shows the effectiveness of the proposed global pose trajectory estimation method.

\begin{figure}[hbt!]
\begin{center}
\includegraphics[width=0.8\linewidth]{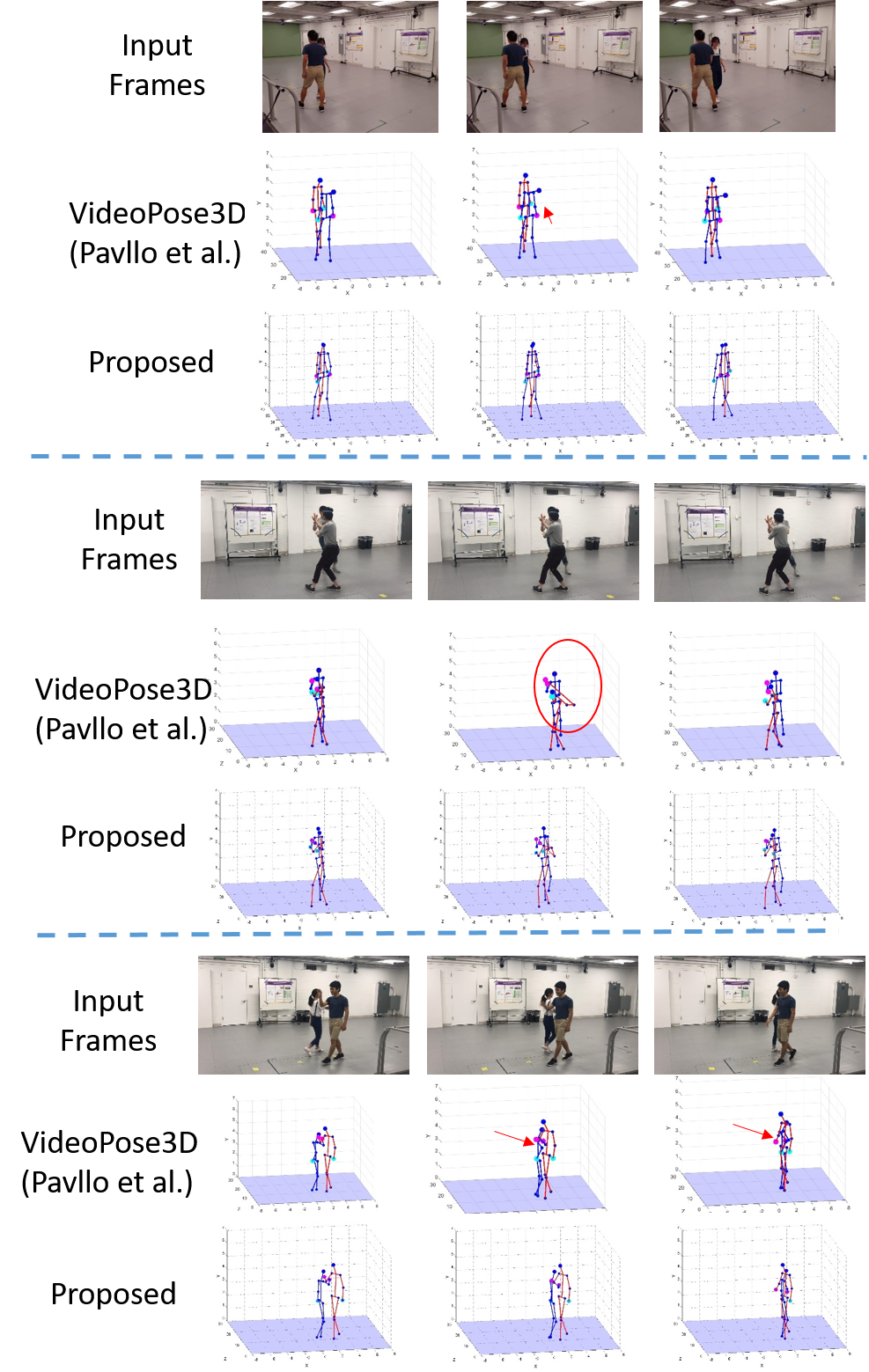}
\end{center}
\caption{Qualitative results on MMHuman. }
\label{fig:qualmmhuman} 
\end{figure}

\hfill
\hfill
\hfill \break
\hfill \break
\begin{table*}[t]
\vspace{1em}
\captionsetup{font=footnotesize}
\begin{center}
\caption{3D Pose Error (Protocol 2, Errors in mm)}
\label{tab:prot2}
\begin{tabular}{L{3cm}C{0.9cm}C{0.9cm}C{0.9cm}C{0.9cm}C{0.9cm}C{0.9cm}C{0.9cm}C{0.9cm}}
\hline\noalign{\smallskip}
& Dir. & Disc. & Eat & Greet & Phone & Photo & Pose & Purch.\\
\noalign{\smallskip}
\hline
\noalign{\smallskip}
Pavllo \textit{et al}. \cite{pavllo20193d} & 70.9 & 92.3 & 69.0 & 107.9 & 66.7 & 98.1 & 69.8 & 73.8 \\
% Proposed & \textbf{56.6} & \textbf{64.2} & \textbf{59.8} & \textbf{101.5} & \textbf{62.6} & \textbf{82.8} & \textbf{62.4} & \textbf{64.0}\\

Proposed & \textbf{53.5}&\textbf{60.3}&\textbf{59.9}&\textbf{87.2}&\textbf{66.7}&\textbf{79.1}&\textbf{59.1}&\textbf{64.8}\\

\noalign{\smallskip}
\hline
\noalign{\smallskip}
& Sit & SitD. & Smoke & Wait & WalkD. & Walk & WalkT. & Avg.\\
\noalign{\smallskip}
\hline
\noalign{\smallskip}
Pavllo \textit{et al}. \cite{pavllo20193d} & 68.1 & 115.4 & 69.7 & 93.8 & 89.8 & \textbf{87.5} & \textbf{77.9} & 83.4\\
% Proposed & \textbf{61.5} & \textbf{101.7} & \textbf{60.9} & \textbf{76.8} & \textbf{80.6} & 105.9 & 91.3 & \textbf{75.5}\\

Proposed&\textbf{63.9}&\textbf{99.7}&\textbf{57.2}&\textbf{73.1}&\textbf{85.5}&107.5 & 80.0&\textbf{73.2}\\

\hline
\end{tabular}
\end{center}
\end{table*}

\begin{table*}[t]
\vspace{1em}
\captionsetup{font=footnotesize}
\begin{center}
\caption{3D Pose Error on MMHuman (Errors in mm)}
\label{tab:errmmhuman}
\begin{tabular}{L{2.5cm}C{1.2cm}C{1.2cm}C{1.2cm}C{1.2cm}C{1.2cm}C{1.2cm}C{1.2cm}}
\hline\noalign{\smallskip}
& ShakeH. & WalkC. & HighF. & PullUp & HandO. & Kungfu & Avg.\\
\noalign{\smallskip}
\hline
\noalign{\smallskip}
Hossain \textit{et al}. \cite{rayat2018exploiting} & 119.3 &119.1 & 119.1 & 120.2 & 115.2 & 119.3 & 118.5 \\
Pavllo \textit{et al}. \cite{pavllo20193d} & 87.7 & 86.1 & 82.8 & 88.8 & 79.1 & 78.7 & 84.2\\
Lin \textit{et al}. \cite{lin2019trajectory} & 110.5 & 116.3 & 123.2 & 116.3 & 107.7 & 118.4 & 114.9\\
Proposed & \textbf{83.7} & \textbf{74.8} & \textbf{78.6} & \textbf{83.9} & \textbf{75.1} & \textbf{76.7} & \textbf{78.3}\\

% \begin{tabular}{L{2.2cm}C{1.1cm}C{1.1cm}C{1.1cm}C{1.1cm}}
% \hline\noalign{\smallskip}
% & ShakeHand & WalkCross & HighFive & - \\
% \noalign{\smallskip}
% \hline
% \noalign{\smallskip}
% Hossain \textit{et al}. \cite{rayat2018exploiting} & 119.3 & 119.1 & 119.1 & - \\
% Pavllo \textit{et al}. \cite{pavllo20193d} & 87.7 & 86.1 & 82.8 & - \\
% Lin \textit{et al}. \cite{lin2019trajectory} & 110.5 & 116.3 & 123.2 & - \\
% Proposed & \textbf{83.7} & \textbf{74.8} & \textbf{78.6} & - \\
% \hline\noalign{\smallskip}
% & PullUp & HandOver & Kungfu & Avg.\\
% \noalign{\smallskip}
% \hline
% \noalign{\smallskip}
% Hossain \textit{et al}. \cite{rayat2018exploiting} & 120.2 & 115.2 & 119.3 & 118.5 \\
% Pavllo \textit{et al}. \cite{pavllo20193d} & 88.8 & 79.1 & 78.7 & 84.2\\
% Lin \textit{et al}. \cite{lin2019trajectory} & 116.3 & 107.7 & 118.4 & 114.9\\
% Proposed & \textbf{83.9} & \textbf{75.1} & \textbf{76.7} & \textbf{78.3}\\
\hline
\end{tabular}
\end{center}
\end{table*}

Table \ref{tab:errmmhuman} shows the results on our recorded dataset MMHuman. This dataset, as explained in Section \ref{sec:dataset}, includes more inter-person occlusions during human interactions. For this dataset, each detected 2D keypoint by OpenPose is associated with a confidence score. We treat it as occlusion if the confidence below 0.3. To test the generalization of the methods, we use the pre-trained model on Human3.6M and not fine-tuned on MMHuman dataset. From Table \ref{tab:errmmhuman}, we can see that our proposed method achieves the best performance. It further proves that our pose estimation method is more robust facing diverse real-world scenarios.

\subsection{Qualitative Results for Occlusion Handling}
Fig.~\ref{fig:qualmmhuman} shows some qualitative results of our proposed method. The recovered 3D poses using our proposed method as well as state-of-the-art method VideoPose3D \cite{pavllo20193d} are displayed in the camera coordinate.  Each example has either severe partial occlusion or full-body occlusion, which is a common problem in human pose estimation the proposed method is targeted at. The first sequence (belonging to "PullUp") shows that being at the end of a inter-person occlusion period, our proposed method demonstrates reliable performance while \cite{pavllo20193d} gets adversely affected. The second sequence ("Kungfu") shows under severe inter-person occlusion, the proposed method still outputs reasonable result, whereas \cite{pavllo20193d} messes up. For the third (”HighFive”) sequence, \cite{pavllo20193d} gives unnatural limb estimations, whereas the proposed method predicts natural poses that are consistent with temporal context. It demonstrates that the proposed method is clearly more reliable than other competing methods in occlusion cases.

\section{Conclusion}
\label{sec:conclusion}
In this paper, we propose a temporal gated convolution model for 3D human pose estimation to address the occlusion issues in the real-world scenario. We also introduce a new 3D human pose dataset MMHuman that features multi-person heavy occlusions and moving cameras, which facilitates evaluation and future research in estimating 3D human poses in real-world scenarios.  Meanwhile, the global pose trajectory is efficiently and effectively estimated via temporal back-projection between 2D and 3D joint sequences. We outperform several state-of-the-art 3D pose estimation methods with temporal models on Human3.6M and our self-recorded dataset MMHuman. For future work, we will combine the recent adversarial learning models to enhance the robustness of the 3D pose reconstruction with occlusion.


% \section*{Acknowledgment}




%\bibliographystyle{IEEEtran}


% \end{thebibliography}

\bibliographystyle{IEEEtran}
\bibliography{egbib}
\end{document}